\documentclass[runningheads]{llncs}
\usepackage[T1]{fontenc}
\usepackage{graphicx}
\usepackage{rotating}
\usepackage{pdflscape}
\usepackage{hyperref}
\usepackage{xcolor}

\begin{document}
%
%
\title{To Pretrain or not to Pretrain? A Case Study of Domain-Specific Pretraining for Semantic Segmentation in Histopathology}
\titlerunning{To pretrain or Not to pretrain? Domain-Specific pretraining histopathology}
%
\author{Tushar Kataria\inst{1,2,4}\and Beatrice Knudsen\inst{3} \and Shireen Elhabian \inst{1,2,4} }

\authorrunning{T. Kataria et al.}
%
\institute{Kahlert School of Computing, University Of Utah \and
Scientific Computing and Imaging Institute, University of Utah \and Department of Pathology, University of Utah \and
corresponding author \\
\{tushar.kataria,shireen\}@sci.utah.edu, beatrice.knudsen@path.utah.edu}

\maketitle    
\begin{abstract}
Annotating medical imaging datasets is costly, so fine-tuning (or transfer learning) is the most effective method for digital pathology vision applications such as disease classification and semantic segmentation. However, due to texture bias in models trained on real-world images, transfer learning for histopathology applications might result in underperforming models, which necessitates the need for using unlabeled histopathology data and self-supervised methods to discover domain-specific characteristics. Here, we tested the premise that histopathology-specific pretrained models provide better initializations for pathology vision tasks, i.e., gland and cell segmentation. In this study, we compare the performance of gland and cell segmentation tasks with histopathology domain-specific and non-domain-specific (real-world images) pretrained weights. Moreover, we investigate the dataset size at which domain-specific pretraining produces significant gains in performance. In addition, we investigated whether domain-specific initialization improves the effectiveness of out-of-distribution testing on distinct datasets but the same task. The results indicate that performance gain using domain-specific pretrained weights depends on both the task and the size of the training dataset. In instances with limited dataset sizes, a significant improvement in gland segmentation performance was also observed, whereas models trained on cell segmentation datasets exhibit no improvement. \href{https://github.com/tushaarkataria/Histopathology-Domain-Specific-Pretraining}{Github Code Repository}.

\end{abstract}

\section{Introduction}

Deep learning models typically require a substantial amount of data to effectively learn generalized latent space representations \cite{deng2009imagenet}. However, acquiring large medical image datasets is more challenging compared to real-world image datasets for three primary reasons. Firstly, the annotation process for medical images involves domain-specific knowledge from pathologists and radiologists to manually outline anatomical structures. This is challenging given the global scarcity of pathology and radiology experts  \cite{metter2019trends,jajosky2018fewer,robboy2020reevaluation}; Secondly, the image annotation interfaces are inefficient in generating labor-intensive workflows. Thirdly, inter-observer disagreement among medical professionals necessitates the involvement of multiple  experts to repeat each annotation task \cite{eaden2001inter,farmer2000importance,nir2018automatic}.
Lastly, in addition to the annotation challenges there are biases in medical data. Biases in histopathology images arise from variations in tissue quality, staining protocols leading to difference in color and texture \cite{howard2021impact}, scanning protocols and slide scanners \cite{howard2021impact,liu2020ms,kataria2023unsupervised}. 

These biases are often site-specific and can cause  major distribution shifts between different data sets, which in turn reduces the generalization of deep learning models \cite{howard2021impact,liu2020ms}.

Other forms of distribution shifts for histopathology in cancer cohorts include discrepancies between cancer and normal tissue histology, the proportion of histologic cancer subtypes, grades and stages, and variations in clinical, demographic, and race-related variables \cite{shi2023ebhi,kataria2023automating}. These variables generate data imbalances that can degrade the performance of deep learning models during testing.

\begin{figure}[!htb]
    \centering
    \includegraphics[scale=0.37]{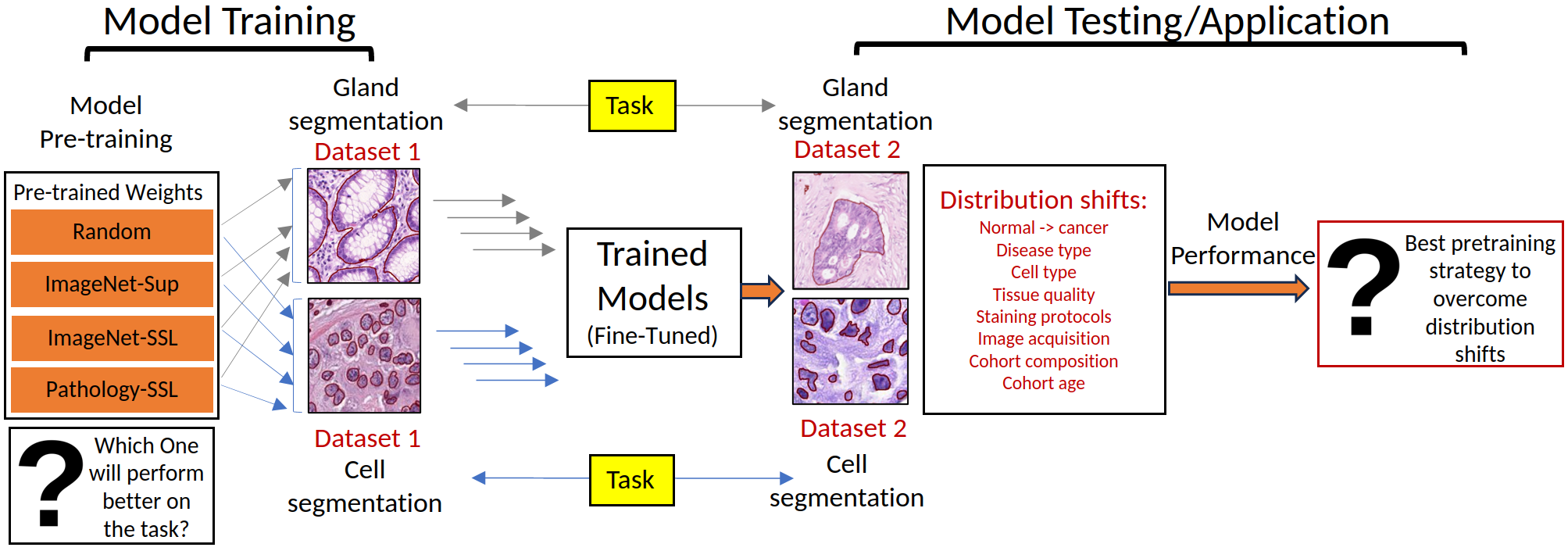}
    \caption{\textbf{Do different weight initialization matter?} The study is designed from the perspective of an AI user who can choose between multiple pretrained models (domain or non-domain, supervised or self-supervised) options for a given task. The best pretrained model is the one that has the highest accuracy on the task and is least affected by distribution shifts. This study provides a framework to choose amongst pretrained models and select the most advantageous for the task.}
    \label{fig:flow_diagram}
\end{figure}

In medical image vision tasks, fine-tuning pretrained models (also known as transfer learning) has become a common approach \cite{erhan2010does,mensink2021factors}. These tasks are important for automated diagnosis, cancer grading and predictions of patients outcomes across all cancer types. 

Using supervised or self-supervised methods, deep learning models exhibit strong capabilities to learn effective latent representations \cite{chen2020big}. However, they may suffer from domain-specific texture bias \cite{hermann2020origins}, which can impede their performance \cite{raghu2019transfusion}.
Previous research indicates that if sufficient data is available for training, a model trained de-novo (i.e., from scratch) may outperform a fine-tuned model \cite{mensink2021factors,raghu2019transfusion}. 
This suggests a potential benefit of domain-specific pretraining \cite{zhang2023large,kang2023benchmarking} over transfer learning from ImageNet \cite{he2016deep}.

Because large, annotated data sets are difficult to obtain for pretraining on histopathology images, self-supervised (SSL) and annotation free methods  provide an alternative strategy for pretraining models to learn valid representation in the latent space \cite{chen2020simple,zbontar2021barlow,caron2020unsupervised}. Models can then be further fine-tuned with a few annotations to produce acceptable results on test datasets. However, no studies systematically evaluated the impact of domain-specific pretaining for histopathology models that are tasked to learn cell and gland segmentation. The closest matching work to this study is an investigation of pretraining on classification and instance segmentation tasks \cite{kang2023benchmarking}.

As gland and cell segmentation differ from instance segmentation and classification, the effect of pretraining on the analysis of out-of-distribution (OOD) datasets also remains unknown. The contributions of this paper are as follows:-
\begin{itemize}
    \item Comparison of de-novo trained models with pretrained models on the ImageNet dataset using class supervision \cite{he2016deep} and self-supervision \cite{zbontar2021barlow} for semantic segmentation tasks in histopathology. 
    \item Finetuning pretrained domain-specific (histopathology) models \cite{kang2023benchmarking} for gland and cell segmentation. These comparisons will indicate whether domain-specific pretraining aids cell and gland segmentation. 
    \item Determining the effect of compute resources and data quantity on model performance improvements.
    \item Investigating  whether  domain-specific  training  leads  to  a  better generalization of models to data distribution shifts.
\end{itemize}
\section{Different Pretrained Weights Used}

To investigate whether domain-specific pretraining leads to generalization in gland and cell segmentation tasks, the study aims to address the following research questions: 
\begin{enumerate}
    \item[$-$] \textit{Is domain pretraining, which involves initializing the weights trained with domain-specific images(histopathology), more effective for transfer learning compared to pretrained weights from ImageNet data?}
    \item[$-$] \textit{Do self-supervised outperform supervised weight initializations?}
    \item[$-$] \textit{Does domain-specific pretraining enhance the quality of features and improve the model's performance on datasets with distribution shifts?}
\end{enumerate}

All initializations are compared against random initialization (i.e., training from scratch), which serves as the baseline to identify initializations (mentioned below) that outperform random. The flow diagram of the study is shown in Figure \ref{fig:flow_diagram}.

Models are trained with 3 different types of initializations: (1) \textbf{pretrained weights using class supervision on ImageNet data}: default weights are provided in Pytorch for \textit{ImageNetV1}  and  \textit{ImageNetV2}. The top-1  accuracies  in  the  initialization  amount to  76.13  and  80.85,  respectively. These weights are obtained by training a ResNet50 \cite{he2016deep} model with class supervision. 
For two other initialization, weights are obtained using a self-supervised technique called Barlow Twins \cite{zbontar2021barlow}. (2) \textbf{Pretrained weights with ImageNet data using SSL (SSLImage):} Self-supervised weights were obtained after training on data from ImageNet without using labels. 
(3) \textbf{Histopathology Domain-Specific pretraining using SSL (SSLPathology):} This model is released as part of the study in \cite{kang2023benchmarking} for domain-specific pretraining on histopathology data. The model was pretrained using more than three million histopathology image patches sampled from various cancers at different magnifications. More details about the pretraining method and the dataset can be found in  \cite{kang2023benchmarking}.

\subsection{Dataset Details}

We have experimented with gland and cell segmentation tasks on these five histopathology datasets:

\textbf{Gland Segmentation Datasets: } Colon cancer datasets, GlaS and CRAG \cite{sirinukunwattana2017gland,graham2019mild}, possess  ground  truth  gland  segmentation  annotations  for normal  and  colon  cancer  glands. The GlaS dataset has 88 training \& 80 testing images of size less 700x600 pixels, whereas the CRAG dataset has 160 training \& 40 testing images of size 1512x1512 pixels.

\textbf{Cell Segmentation Datasets: }
Three cell segmentation datasets are used for experimentation KUMAR \cite{kumar2017dataset}, CPM17 \cite{vu2019methods} and TNBC \cite{naylor2018segmentation} possess  ground  truth  annotations  of  nuclear outlines.
\begin{table}
\caption{\textbf{Cell Segmentation Dataset Details.} Sample examples of the dataset are shown in supplementary Figure \ref{fig:cell_segmentation_results}.}
    \centering
    \begin{tabular}{l|c|c|c|c}
       Datasets & Train Imgs& Test Imgs& Img Size &  No. of Annotated Nuclei \\
       \hline
       KUMAR \cite{kumar2017dataset}  & 16 & 14 &1000x1000& 21623 \\
       CPM17 \cite{vu2019methods}  &  31 & 31 & 500x500  & 7570\\
       TNBC \cite{naylor2018segmentation} & 34& 16 & 512x512 & 4022\\
    \end{tabular}
    \label{tab:my_label}
\end{table}
\vspace{-0.2in}

\subsection{Implementation Details}
A U-Net \cite{ronneberger2015u} model is used with \textit{Resnet50} \cite{he2016deep} backbone for semantic segmentation application(gland \& cell both). The decoder is always the same for all models. Details of implementation can be found in the associated github\footnote{\href{https://github.com/tushaarkataria/Histopathology-Domain-Specific-Pretraining}{GitHub Repository}}.   Models are trained using  PyTorch  and  a data  split  of  80-20 for  training and  validation.  The best  model  possessing  a  minimum  loss  on  validation  data  is further evaluated on the test dataset. Testing data is only used for inference.

During training the patch size is 256x256, sampled randomly in the whole image. At inference, predictions are averaged over a window size of 128 pixels. The learning rate is fixed to 0.0001 and the number of epochs for all experiments is set to 4000 for gland segmentation and 2000 for cell segmentation. The models are trained five times and average metrics are reported, this ensures that variations due to stochasticity caused by the dataset  loader are factored out. Data  augmentation  includes  horizontal and  vertical  flips,  random  rotation,  and  translation.  All models  are  trained  on NVIDIA V100 GPUs. 

\textbf{Evaluation Metrics:}
Dice  and  Jaccard  scores  (also  known  as  the intersection over union) serve as metrics for segmentation tasks   \cite{sirinukunwattana2017gland,ronneberger2015u}.

\section{Results}

\begin{figure}[!htb]
    \centering
    \includegraphics[scale=0.38]{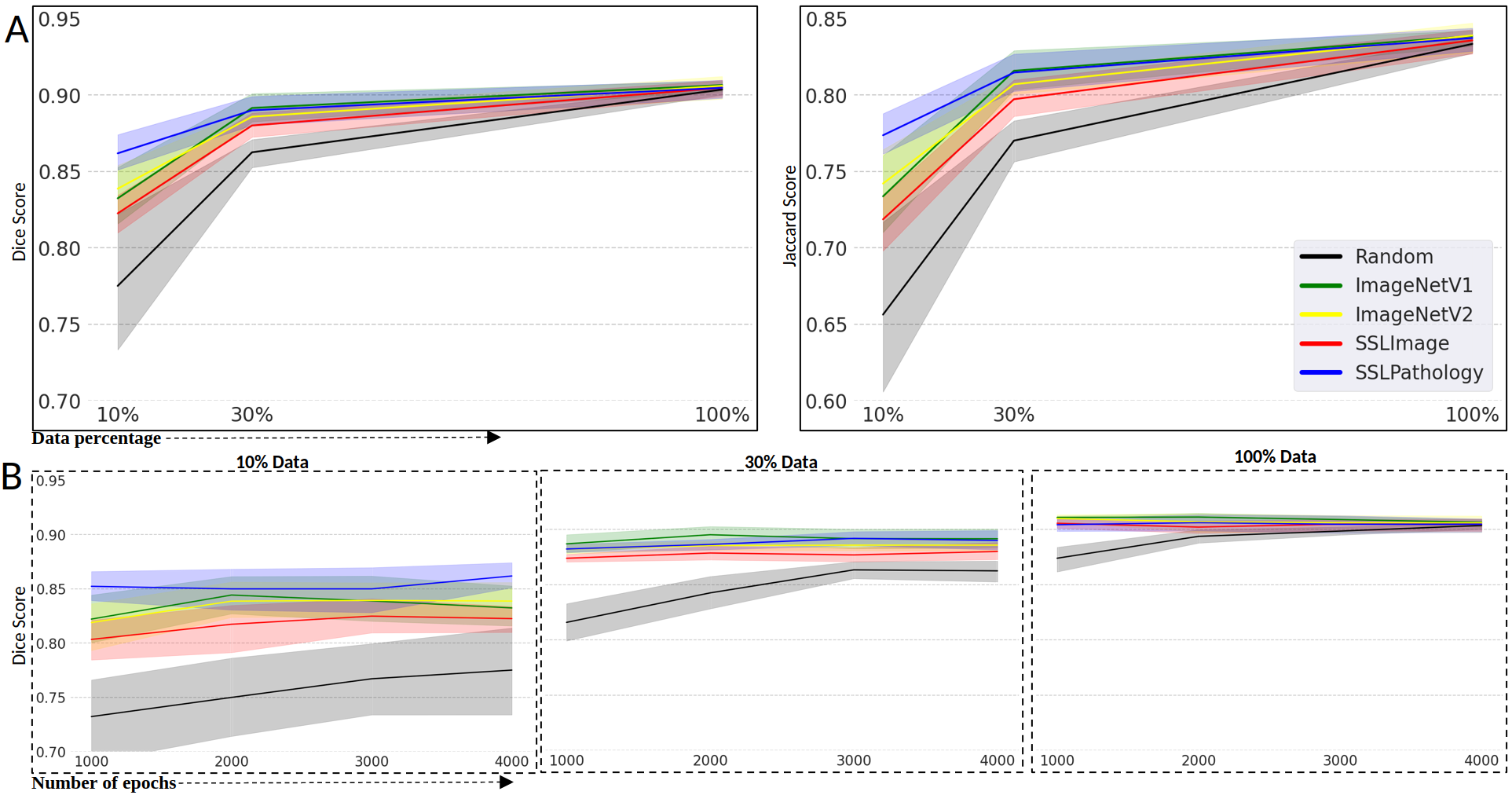}
    \caption{\textbf{Gland Segmentation Results for Different Initializations on GlaS\cite{sirinukunwattana2017gland}.} (\textbf{A}) Dice and Jaccard Score for different percentage of training data used. Solid line shows the mean performance metrics, where as color shading shows the variance. We can clearly observe that increasing data increases model performance, but with more data domain specific pretraining doesn't have a significant effect on performance. (\textbf{B}) Average dice score variations with different amounts of training time, i.e., number of epochs. We clearly see a difference in performance for different initialization for low dataset size and lesser epochs. Results on CRAG dataset are shown in Supplementary Figure \ref{fig:my_label:gland_segmentation_Crag}.}
    \label{fig:training_dataset_size_gland}
\end{figure}

\begin{figure}[!htb]
    \centering
    \includegraphics[scale=0.35]{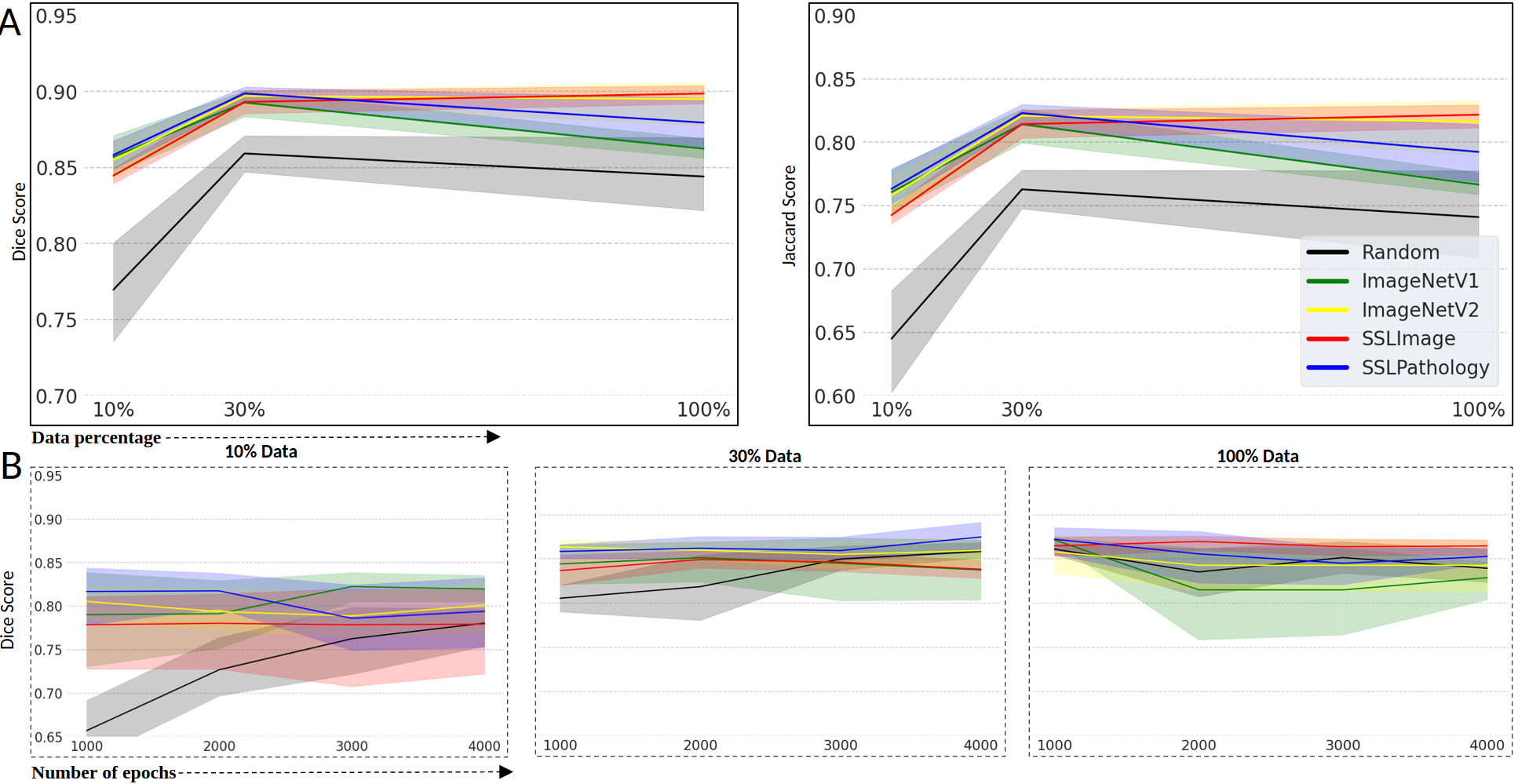}
    \caption{\textbf{Gland Segmentation Results for Different Initializations on CRAG\cite{graham2019mild}.} (\textbf{A}) Dice and Jaccard Score for different percentage of training data used. (\textbf{B}) Average dice score variations with different amounts of training time, i.e., number of epochs.} 
    \label{fig:my_label:gland_segmentation_Crag}
\end{figure}

\textbf{Gland Segmentation Results:} The line plots (standard deviation marked as shading) of performance measures for different initialization are shown in Figure \ref{fig:training_dataset_size_gland}-A and \ref{fig:my_label:gland_segmentation_Crag}-A. We  trained  models  with different backbone  initializations  on  an  increasing  amount  of  data. The  following  observations  emerged  from  these  experiments:- (a) Increasing the quantity of data improves performance for all initializations and decreases variation. (b) For different training dataset sizes, models with pretrained weight initializations outperform those with random initializations, but the performance gap between models trained with random initialization and pretrained weights decreases as the quantity of data increases. 
(c) For small datasets, finetuning domain-specific pretrained weights has a significant performance advantage over other initializations. However, as dataset size increases, the effect diminishes \footnote{All images are best viewed on a digital device following magnification.}. 

Variations in performance due to different amounts of training epochs  for all datasets are shown in Figure \ref{fig:training_dataset_size_gland}-B and \ref{fig:my_label:gland_segmentation_Crag}-B. For very small datasets(10\% and 30\% graph), finetuning domain-specific pretrained weights outperforms all other initializations at all epochs. However, for larger datasets(100\% data), ImageNet supervised weights also outperform at lower epochs. This show that domain-specific pretraining is not dependent on computational power but rather on dataset diversity. If a dataset is not diverse or small in size, then domain-specific pretrained weights are beneficial, but other initialization can be better for higher diversity and epochs. Qualitative results are shown in supplementary Figure \ref{fig:gland_segmentation_results}, domain-specific fine-tuned models have more accurate gland outlines and fewer false positive pixels than other models.

\begin{figure}[!htb]
    \centering
    \includegraphics[scale=0.32]{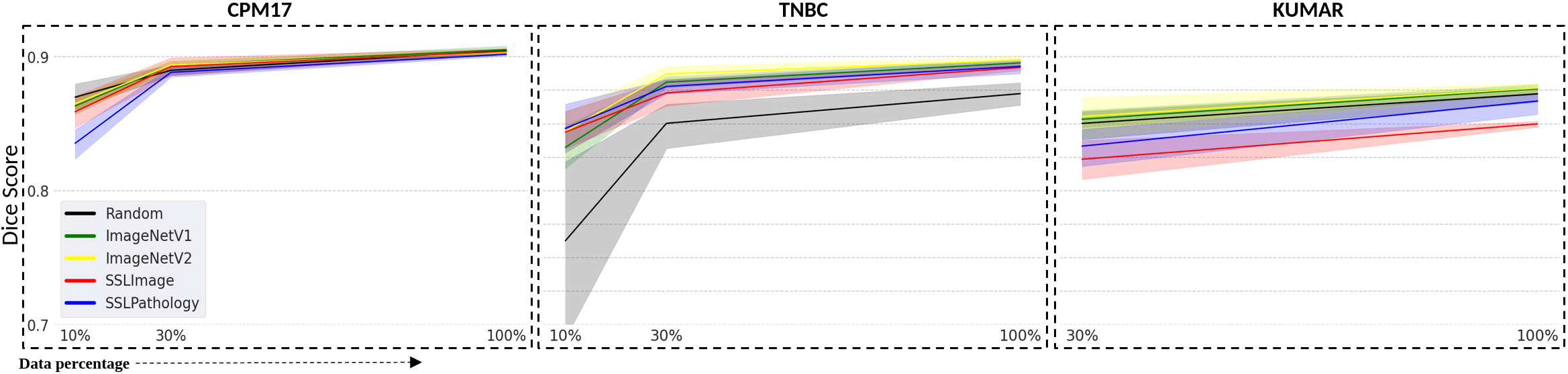}
    \caption{\textbf{Cell Segmentation Results.} Different initialization has similar patterns, i.e., with increasing data variation in performance decreases and mean performance increases. But for cell segmentation domain specific pretraining doesn't seem to be better than image-net pretrained weights for different data sizes.}
    \label{fig:cell_segmentation:dice_jaccard}
\end{figure}

\textbf{Cell Segmentation Results :} The performance of various initializations is depicted in Figure \ref{fig:cell_segmentation:dice_jaccard}. Even though some of the observations are similar to those of previous experiments, novel observations emerge from cell segmentation results:- (a) Model  performances  with KUMAR \cite{kumar2017dataset} data are an exception where random initialization is outperforming or competitive with other initializations. (b) Domain-specific pretraining is performing similar to or worse than ImageNet initialization for most cases. Altogether  our  results  demonstrate  that fine-tuning  domain-specific  pretrained weights  does  not  improve  the  performance  of  the  U-Net/ResNet  model  for  cell segmentation tasks. Qualitative results are shown in supplementary Figure \ref{fig:cell_segmentation_results}.

\begin{figure}[!htb]
    \centering
    \includegraphics[scale=0.38]{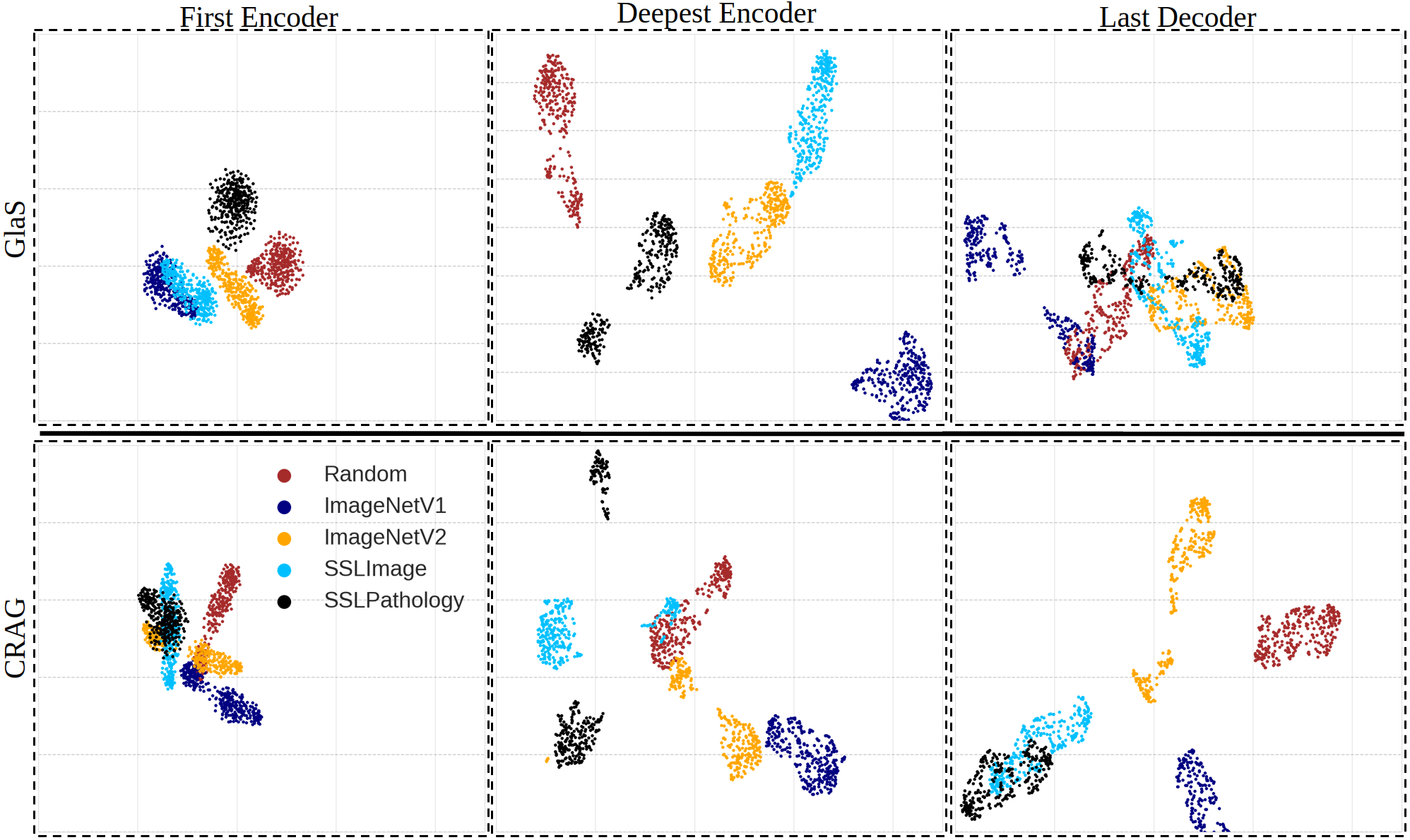}
    \caption{\textbf{UMAP for different Gland Segmentation models for GlaS and CRAG datasets.} We generate UMAP with nearest neighbor=25, distance=0.1, and metric=cosine. Comparing the latent representation of the initial encoder to that of the deepest encoder, there is substantially less overlap between initializations, but the distribution of points is topologically similar.}
    \label{fig:UMAP_results}
\end{figure}

\textbf{UMAP Results:} We sampled 300 random patches from the test sets of GlaS and CRAG to generate projections for encoders and decoders shown in Figure \ref{fig:UMAP_results}. Feature values were extracted from the first encoder layer in U-Net, the deepest encoder layer, and the last decoder layer. 

In the network's first layer, the projections of features from various initializations form clouds that overlap.  We interpret this observation to conclude that the initial layers of deep neural networks capture low-level statistics and that all initializations capture comparable attributes. As encoding depth increases, the representations become more distinct and the overlap decreases, indicating that networks pretrained in different ways may be learning different representations of the same data. This is counterintuitive, as we would expect that each of the pretrained models generates similar high-level representations when performing identical tasks and using the same dataset. However, the distribution of features in the UMAP projection of latent layer representations appears to have topological similarity across initializations which indicates that features for different initialization may be related via a rigid transformation in latent space. A similar conclusion is valid for the decoder UMAP. Together, these results suggest that distinct initializations, despite being clustered at different locations in the UMAP, might learn similar relational feature characteristics between samples in the dataset. 
\subsection{Out Of Distribution Testing Results} 
For OOD testing, we use the fine-tuned segmentation models (source dataset, Dataset 1) with different initialization for out-of-distribution testing on other datasets (target dataset, Dataset 2) without fine-tuning on the second dataset. This analysis reveals the distribution bias various initializations learned during the training (fine-tuning) process.

\begin{figure}[!htb]
    \centering
    \includegraphics[scale=0.29]{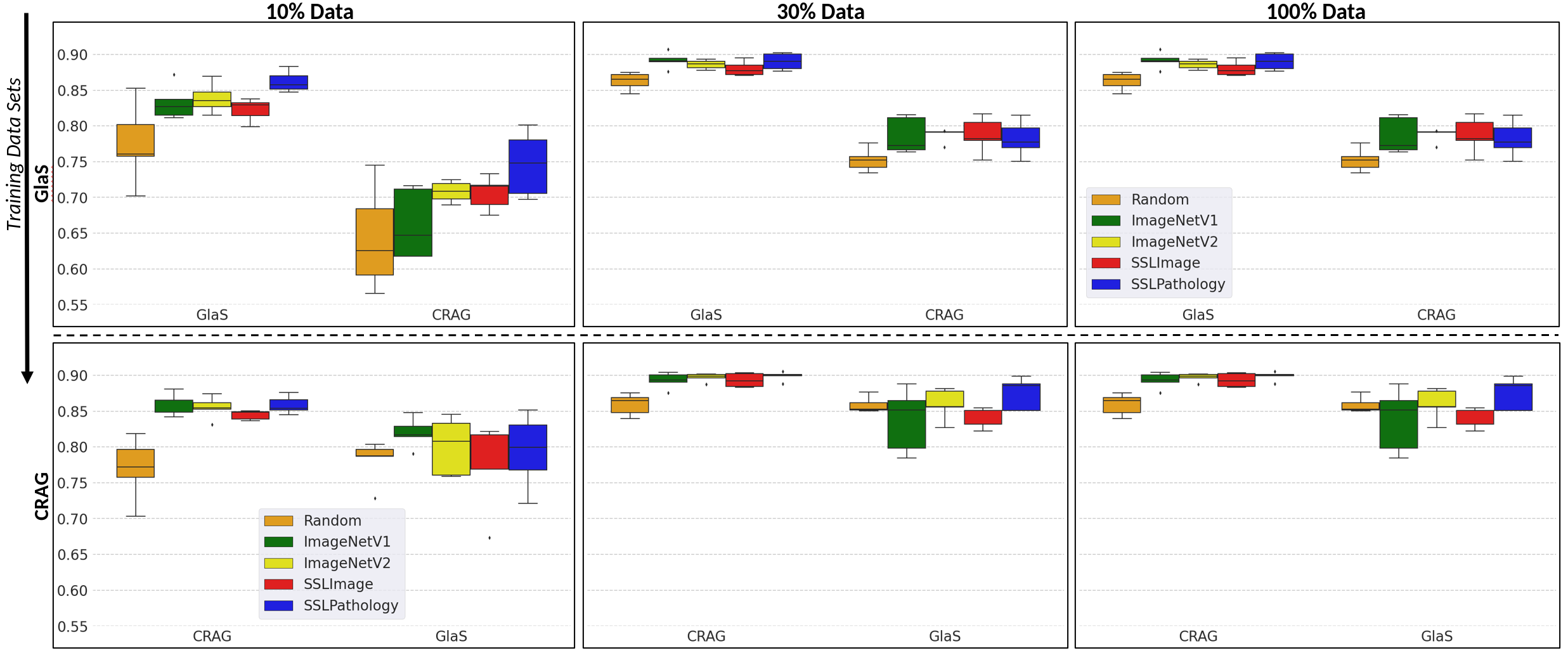}
    \caption{\textbf{Average dice score for OOD testing.} Y-axis shows the dataset used for training of the model. X-axis is the performance on the corresponding test sets without any fine-tuning. A model trained on CRAG datasets transfers effectively to GlaS, but not vice versa. Domain-specific pretrained models are generally better at out-of-domain performance.} 
    \label{fig:my_label:OOD_gland}
\end{figure}

\textbf{Gland Segmentation Results} The results for OOD testing for the gland segmentation task are shown in Figure \ref{fig:my_label:OOD_gland}. At a low amount of data, the domain-specific, finetuned models perform best and using random initializations results in the greatest relative performance drop compared to all other initializations.

\textbf{Cell Segmentation Results:} The results of OOD testing for different datasets are shown in supplementary Figure \ref{fig:my_label:OOD_nuclues} and lead to the following observations: (a) pretrained models are better than models with random initialization at the same task on unseen datasets from  KUMAR \cite{kumar2017dataset} and CPM17 \cite{vu2019methods}). In contrast, models with random initialization and trained on TNBC \cite{naylor2018segmentation} outperform or perform the same as the pretrained initialized model. (b) A drop in performance exists on TNBC data for models trained on KUMAR \cite{kumar2017dataset} and CPM17 \cite{vu2019methods} but not for models trained on TNBC \cite{naylor2018segmentation} or KUMAR \cite{kumar2017dataset} and applied to CPM17. (c) Domain-specific pretrained models when tested on OODdata demonstrate a lesser drop in performance compared to other pretraining approaches.

\section{Conclusion and Future Work}
In this study, we demonstrate that a using domain-specific pretrained backbone can be beneficial for gland segmentation when data are limited or of low diversity data for the task at hand. However, the performance gap for fine-tuned domain-specific pretrained weights compared to random initialization decreased as the amount of training data increased. The results of cell segmentation indicate that domain-specific pretrained weights may not be advantageous for all types of tasks. 
The results of UMAP projections indicate that the initial layers of domain-specific and non-domain-specific models learn similar features, but that the deeper encoders are distinct. Although the topology of latent feature representations is similar for the different initialization, models may be learning similar high-level characteristics within the latent feature spaces. Lastly, during out-of-distribution testing, models initialized with domain-specific pretrained weights suffered the same performance degradation as other initializations. Therefore, domain-specific pretrained weight initialization may not be effective at learning site-independent features. 
Our final conclusion from this study is that fine-tuning domain-specific pretrained weights may be beneficial for specific tasks and datasets, but benefits are not universal. Domain-specific pretrained weights suffer from the same issues as weights on image-net. Lastly, we would like to make the reader aware that this study did not cover medical vision tasks such as multi-class semantic segmentation and cell detection. We also did not utilize models pretrained using vision-language models. Both these comparisons are left for future work.
\section*{Acknowledgments}
We thank the Department of Pathology and the Kahlert School of Computing at the University of Utah for their support of this project. 

 \bibliographystyle{splncs04}
 \bibliography{paper48}
\newpage
\section{Supplementary}
\begin{figure}[!htb]
    \centering
    \includegraphics[scale=0.27]{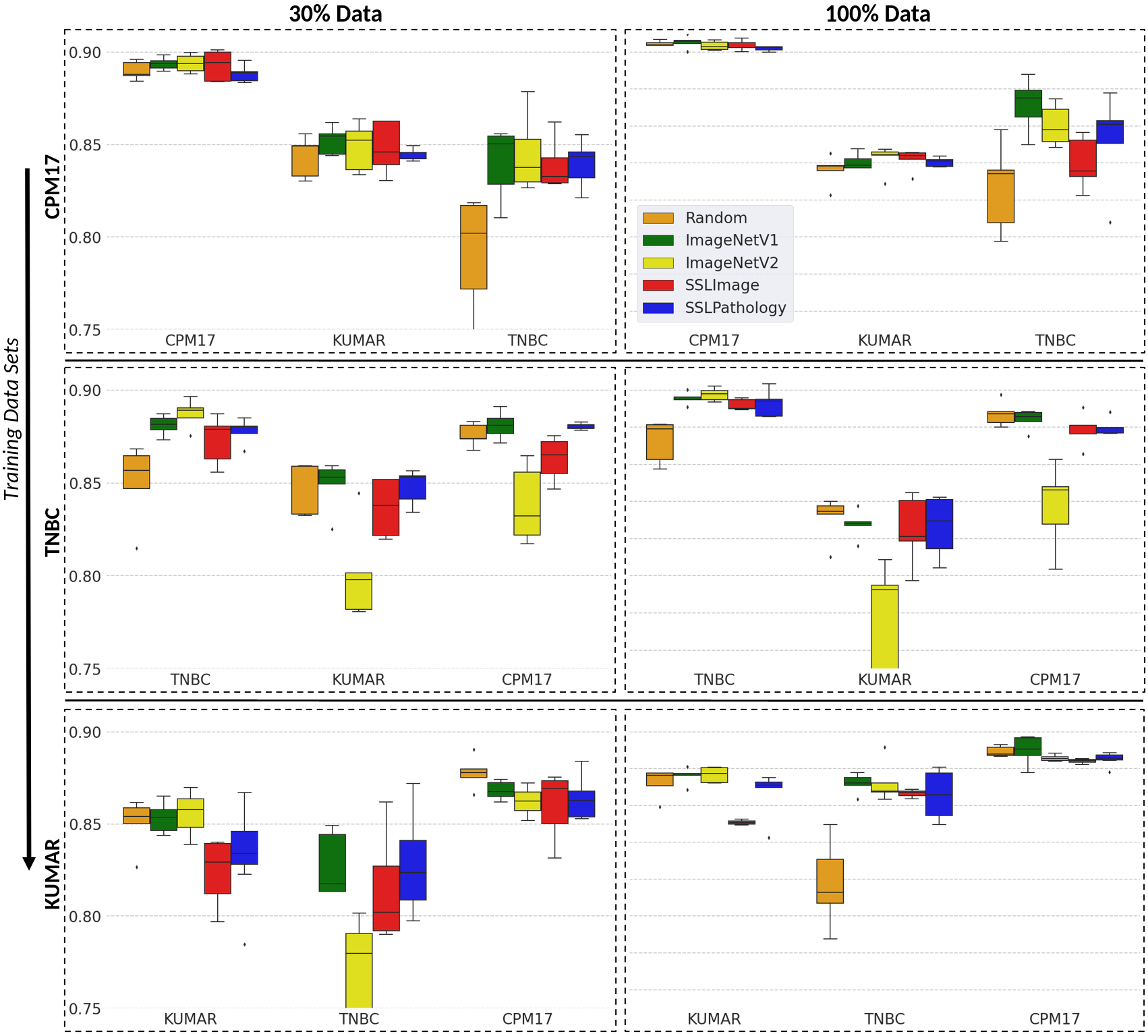}
    \caption{\textbf{Average dice score for OOD testing.} Y-axis shows the dataset used for training the model. X-axis is the performance on the test set of the datasets without any fine-tuning.} 
    \label{fig:my_label:OOD_nuclues}
\end{figure}

\begin{figure}[!htb]
    \centering
    \includegraphics[scale=0.29]{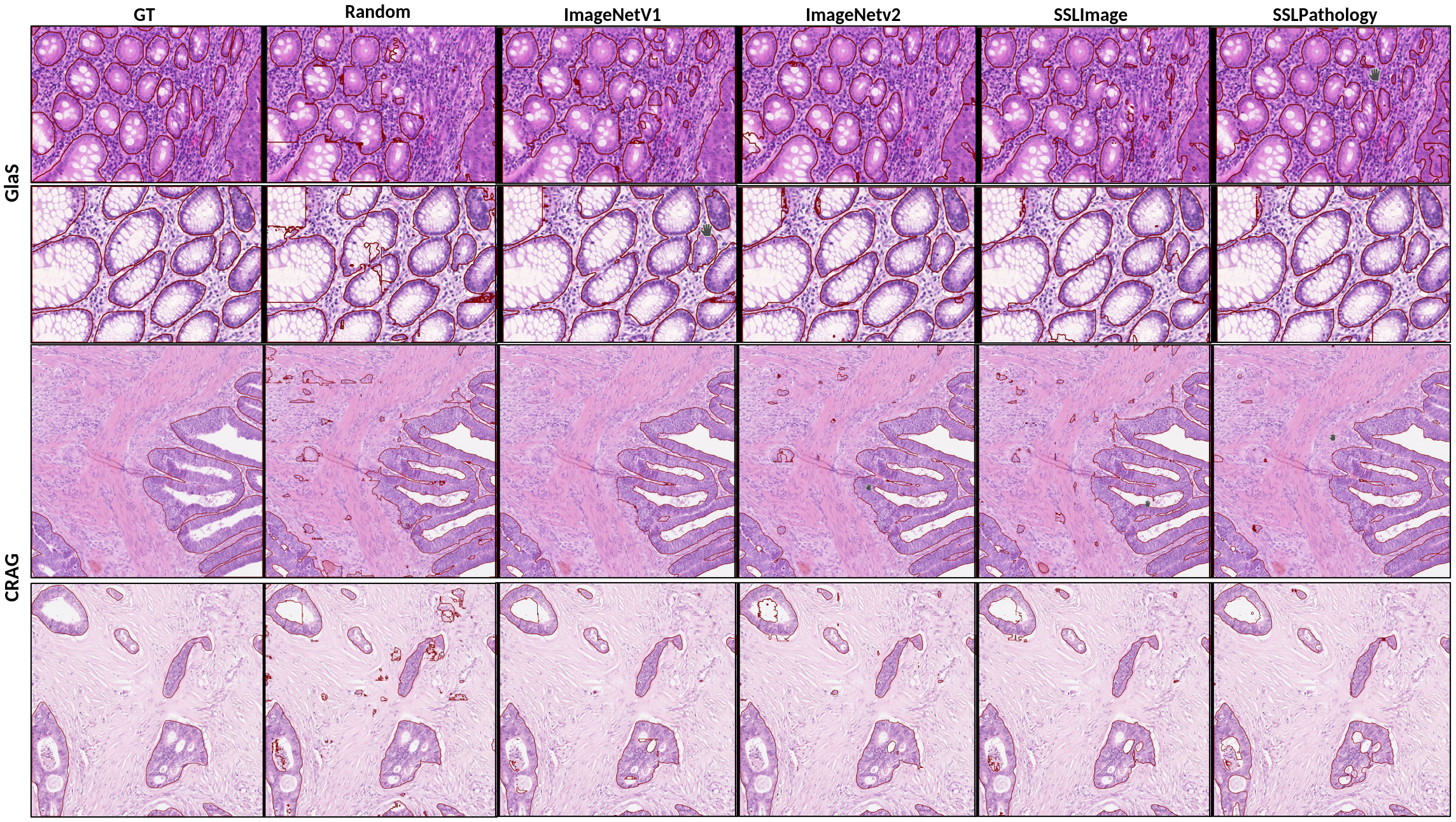}
    \caption{\textbf{Qualitative results for Gland Segmentation Experiments.}  We can observe that pretrained models have better qualitative results than Random Initializations. Domain Specific pretraining models perform better for gland segmentation tasks. These models are better at recognizing the outlines of the gland compared to other initializations. }
    \label{fig:gland_segmentation_results}
\end{figure}
\begin{figure}[!htb]
    \centering
    \includegraphics[scale=0.30]{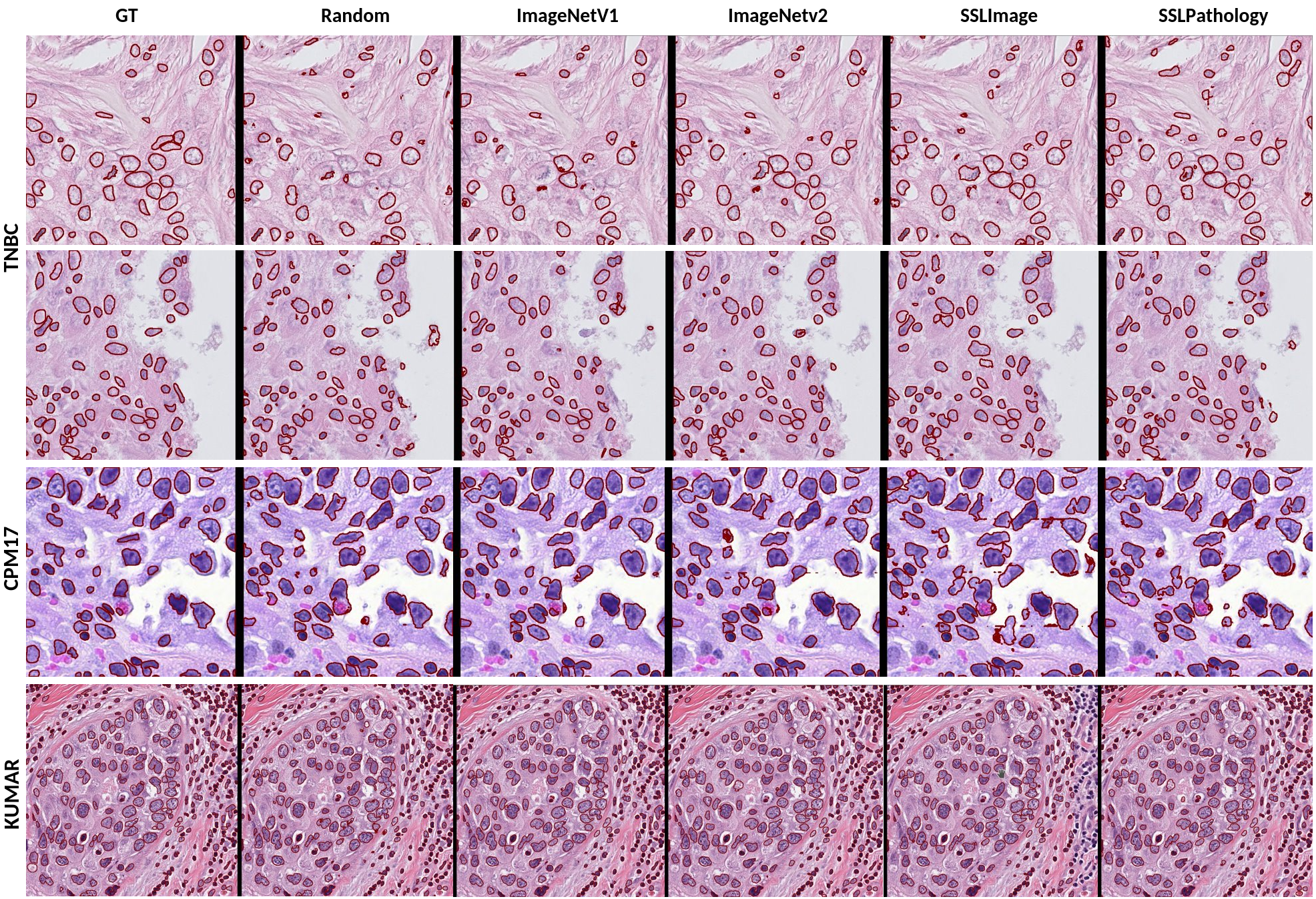}
    \caption{\textbf{Qualitative results for Cell Segmentation Experiments.} We can observe that pretrained models have better qualitative results than Random Initializations. But all of the pretrained models make similar mistakes in the outlines of cells, touching cells and distinguishing between cell and background.}
    \label{fig:cell_segmentation_results}
\end{figure}

\end{document}